# Heterogeneity-Aware Personalized Federated Learning for Industrial Predictive Analytics

Yuhan Hu

*Edward P. Fitts Department of Industrial and Systems Engineering, North Carolina State University, Raleigh, NC, USA*

and

Xiaolei Fang

*Edward P. Fitts Department of Industrial and Systems Engineering, North Carolina State University, Raleigh, NC, USA*

**Abstract**

Federated prognostics enable clients (e.g., companies, factories, and production lines) to collaboratively develop a failure time prediction model while keeping each client's data local and confidential. However, traditional federated models often assume homogeneity in the degradation processes across clients, an assumption that may not hold in many industrial settings. To overcome this, this paper proposes a personalized federated prognostic model designed to accommodate clients with heterogeneous degradation processes, allowing them to build tailored prognostic models. The prognostic model iteratively facilitates the underlying pairwise collaborations between clients with similar degradation patterns, which enhances the performance of personalized federated learning. To estimate parameters jointly using decentralized datasets, we develop a federated parameter estimation algorithm based on proximal gradient descent. The proposed approach addresses the limitations of existing federated prognostic models by simultaneously achieving model personalization, preserving data privacy, and providing comprehensive failure time distributions. The superiority of the proposed model is validated through extensive simulation studies and a case study using the turbofan engine degradation dataset from the NASA repository.

# 1. Introduction

Remaining useful life (RUL) prediction is to estimate the amount of time that a component can function properly before failure, which is essential in industry due to its role in preventing unscheduled downtime and optimizing maintenance schedules [1]. Approaches of RUL prediction can be categorized into model-driven methods and data-driven methods [2]. Model-driven methods employ physics-based models that require expert domain knowledge to build analytical or approximation equations, such as differential equations describing degradation kinetics or fatigue life equations based on stress–strain relationships. In contrast, data-driven methods utilize machine learning algorithms that are learned directly from data. Since data-driven approaches impose fewer requirements on prior knowledge of the underlying failure mechanisms, they can uncover complex patterns, such as sensor correlations, that might be difficult for human analysts to detect. As a result, data-driven methods usually outperform traditional model-driven ones, particularly when degradation processes are complex, and data is high-dimensional [3]. In system prognostics, data-driven approaches are typically achieved by establishing machine learning and statistical models that map condition monitoring sensor signals to their time-to-failure (TTFs) [4-6].

Most existing data-driven approaches assume there is a large amount of historical data, i.e., degradation signals and their corresponding failure times, for model training [7-11]. Unfortunately, the historical failure data owned by an individual client or organization are often very limited, making it difficult to develop an effective prognostic model independently. One naïve solution that allows multiple distributed clients to jointly train a prognostic model is cloud computing-based prognostics. It works by first uploading the data from multiple clients to a centralized cloud server, then training a prognostic model based on the aggregated data on the cloud. The trained model is finally distributed back to the clients for real-time condition monitoring. However, it introduces two primary challenges for industrial systems and end clients. The first challenge is data privacy and security regulations [12-15]. Most data are highly sensitive in nature, and cannot be simply transferred due to privacy protection policies and regulatory compliance. Many organizations are even not willing to share their data with the cloud since it may contain some proprietary information and operational metrics. The second concern is computational and communication resource constraints. Although each client may have only a few failure records, each record contains a large amount of multi-sensor data collected under various operating conditions. Consequently, centralizing such data involves transmitting substantial data volumes to the server, leading to high communication costs.

To overcome these limitations, FL-based prognostics has gained increasing popularity. FL is a distributed learning paradigm that allows multiple data owners, ranging from individual mobile devices to entire organizations, to collectively train their models in a privacy-preserving manner [16]. Numerous current FL training methods originate from the federated averaging (FedAvg) algorithm introduced in [17].

It consists of two steps, including local training and global aggregation. First, clients are facilitated to train machine-learning models with their isolated data locally. Then, instead of transferring the raw data, only the model updates are sent back to the server to be combined as a single global model. Since data is not directly shared during the training process, it effectively addresses common challenges in data privacy and security. Thus, FL-based prognostics is a valuable choice for scenarios where data cannot be centralized due to regulatory constraints. It allows multiple organizations with limited data to jointly train a model for RUL prediction without sharing their raw data, which is widely applied in aviation, automotive, and semiconductors. Moreover, FL is also communication efficient as only the model parameters are transmitted, which are considerably smaller in size than raw data.

Recent studies have made some efforts on developing FL-based prognostic models, however, most existing models assume data homogeneity across clients, meaning that all clients are considered to operate the same type of equipment and components under identical conditions [5, 6, 18-22]. In real-world applications, this assumption often does not hold, as client data are frequently heterogeneous. For example, two clients may utilize different but functionally similar types of bearings. In another case, clients may employ the same bearing type, but the operational conditions vary, for instance, one client's equipment may operate at 1600 rpm, while another's runs at 2000 rpm. In such heterogeneous scenarios, the underlying degradation processes may share similar characteristics but are not strictly identical across clients. Thus, the performance of existing prognostic models is compromised due to the violation of the homogeneity assumption.

Several studies have attempted to address the challenges of client-level heterogeneity by developing methods that explicitly account for variations across clients in federated prognostic modeling [23-29]. For example, Chen et al. [23] employed deep neural networks in an FL framework to learn a shared global health degradation representation alongside client-specific model heads for a personalized RUL prediction. Lu et al. [29] proposed a heterogeneous federated transfer learning method with knowledge distillation for lithium-ion battery RUL prediction, where the server trains device-tailored networks on public data, and devices collaboratively refine them by exchanging distilled insights. However, these methods still suffer from some common limitations. First, almost all of them are based on deep learning, while statistical learning models are being ignored. Deep learning-based prognostic models involve a large number of parameters to be trained, which can be time-consuming and infeasible for edge clients with limited computing power. Besides, deep learning-based methods require a large volume of training data, while collecting sufficient failure data is difficult, even when pooling data from multiple organizations. In contrast, statistical learning models are often more efficient, interpretable, and better suited for scenarios with limited data and restricted computation capacity. Second, current FL-based prognostic models considering heterogeneity can only provide a point estimate of the RUL prediction instead of a failure time distribution.

However, many decision-making tasks, such as maintenance scheduling and inventory optimization, require a distribution of predicted failure times rather than a single point estimate. Without this, integrating proposed models into operational processes remains challenging.

To address these challenges, this study proposes a personalized federated prognostic model designed for clients with heterogeneous data. The proposed approach enables client-specific prognostic models that are structurally similar yet tailored to individual clients. Each client trains its local model on its own data, and borrows useful information from other similar models during training while preserving its unique characteristics. This is achieved by training a (log)-location-scale (LLS) regression model for each client and encouraging the regression coefficients to be similar but not identical across clients. LLS regression is a type of regression model that extends the conventional linear regression model, accommodating the scenarios where the response variable may exhibit skewness or heteroscedasticity [30]. It assumes the response variable follows a distribution from the (log)-location-scale family, such as normal, log-normal, logistics, log-logistics, small extreme value (SEV), and Weibull. When the response variable is sampled from a normal, logistic, or SEV distribution, the model is termed as location-scale regression. When the response variable follows a log-normal, log-logistic, or Weibull distribution, it is termed as log-location-scale regression. Since it covers most of the failure time distributions of engineering assets in industrial applications, this flexibility makes it particularly useful in prognostics. Another key advantage is its ability to provide a distribution of the predicted failure time rather than a point estimate. This feature is particularly valuable for informed decision-making in applications such as predictive maintenance and inventory control. The parameters in LLS regression can be estimated using maximum likelihood estimation (MLE), equivalently, it is solved by minimizing the negative log-likelihood function. To realize the goal that the regression models are similar but not identical across clients, we formulate an optimization problem that has two components: the first is the summation of the loss function of each client, and the second is a regularization term that penalizes the clients whose model parameters are diverse too much from other clients. By solving the two components together, the combined objective eventually coincides with the optimal solution with better generalization across clients.

In order to allow multiple clients to jointly train the proposed FL prognostic models while keeping their data local and confidential, we propose a two-step optimization algorithm, which is solved using proximal gradient descent (PGD). PGD is well-suited for convex optimization problems with decomposable objectives. It operates by alternately taking a gradient step on one part of the objective and applying a proximal operation on the other. In our model, the objective consists of a local loss function that reflects the client's private data and a regularization term that encourages similarity between models of related clients. Thus, the optimization process involves two key steps. First, clients collaboratively update their parameters by optimizing the regularization term on the server through gradient descent. This step involves communication with similar clients via weighted model aggregation between each local model and its corresponding personalized cloud model. The collaborative mechanism gradually aligns model parameters among clients with similar degradation patterns. Second, each client performs a proximal update based on its local loss function, allowing further refinement of the model using private data. The two-step PGD framework enables personalized training in a federated setting while maintaining privacy and addressing data heterogeneity across clients.

The remainder of the paper is organized as follows. Section 2 presents the development of the proposed federated prognostic model. Section 3 introduces the federated parameter estimation algorithm based on PGD, where we first outline the general steps of PGD, and detail its implementation within our personalized learning framework. Sections 4 and 5 evaluate the effectiveness of the proposed model using simulated data. A case study with a degradation dataset from the NASA data repository is provided in Section 6. Finally, Section 7 concludes this study.

## 2. Federated Prognostic Model Development

In this section, we will introduce the development of the proposed federated prognostic model, which is designed to address data heterogeneity across distributed clients. In practical industrial scenarios, each client operates machines under distinct environmental and loading conditions, leading to degradation processes that are similar in nature but not identical. Relying on a single shared global model to represent all clients often fails to capture their local variations, which results in suboptimal prediction performance. On the other hand, training independent local models preserves client-specific characteristics but suffers from limited data availability, making it difficult to achieve reliable RUL predictions under diverse operating conditions. To overcome these, this article proposes a personalized federated prognostic model that can be collaboratively constructed by multiple heterogeneous clients, allowing each client to develop a tailored prognostic model by leveraging information from similar clients while preserving data privacy. In the following paragraphs, we will first introduce the formulation of each client's local prognostic model, and then introduce an optimization problem that allows all clients to jointly estimate their model parameters in a privacy-preserving manner.

For each client $i$, the prognostic model is commonly developed using functional LLS regression, which maps the TTF of a system against its multi-stream degradation signals. Due to the complexity of parameter estimation, in this paper, we transform the functional LLS regression into classic LLS regression, which has proved an equivalence in terms of information retention and model performance in [1]. Details of the proof and steps can be found in [1, 5]. Thus, the high-dimensional degradation signals of the $j^{th}$ system of client $i$ are transformed into low-dimensional features, $\{x_{ijk}\}_{k=1}^{K}$, where $j = 1, \ldots, n_i$. Accordingly, the classic LLS regression is in the form of:

$$y_{ij} = \boldsymbol{\beta}_i^T \boldsymbol{x}_{ij} + \sigma_{ij}\epsilon \tag{1}$$

where $y_{ij} \in \mathbb{R}$ is the TTF of the system $j$ from client $i$, $\boldsymbol{x}_{ij} = (1, x_{ij1}, \ldots, x_{ijK})^T \in \mathbb{R}^{K+1}$ is the vector of extracted features for the $j^{th}$ system of client $i$, $\boldsymbol{\beta}_i = (\beta_{i0}, \beta_{i1}, \ldots, \beta_{iK})^T \in \mathbb{R}^{K+1}$ is the coefficient vector of client $i$, $\sigma_{ij}$ is the scale parameter, and $\epsilon$ is the random noise term with a standard location-scale density $f(\epsilon)$. The probability density function of $y_{ij}$ is expressed as:

$$f_{Y_{ij}}(y_{ij}) = \frac{1}{\sigma} f\left(\frac{y_{ij} - \boldsymbol{x}_{ij}^T \boldsymbol{\beta}_i}{\sigma}\right) \tag{2}$$

where $f(\cdot)$ is the standard location-scale density function. For example, $f(\epsilon) = 1/\sqrt{2\pi}\exp(-\epsilon^2)$ if $\epsilon$ is sampled from a normal distribution, and $f(\epsilon) = \exp(\epsilon - \exp(\epsilon))$ for SEV distribution.

To estimate the parameters in Eq. (1), a straightforward solution is to use MLE for each client independently. However, this approach often suffers from limited data availability and neglects the relationship among clients, which leads to poor generalization in heterogeneous scenarios. To let the models across clients be similar but not identical, we formulate a joint optimization problem that enables $m$ clients to collaboratively estimate their parameters in Eq. (1), including $\beta_i$ and $\sigma_i$, where $i = 1, \ldots, m$. The joint formulation integrates two components, which are given as:

$$\min_{\boldsymbol{B},\boldsymbol{\sigma}}(L(\boldsymbol{B},\boldsymbol{\sigma}) + \lambda G(\boldsymbol{B},\boldsymbol{\sigma})) \tag{3}$$

where $\boldsymbol{B} = [\boldsymbol{\beta}_1, \boldsymbol{\beta}_2, \ldots, \boldsymbol{\beta}_m] \in \mathbb{R}^{(K+1)\times m}$ collects the coefficients $\boldsymbol{\beta}_1, \boldsymbol{\beta}_2, \ldots, \boldsymbol{\beta}_m$ as its columns, $\boldsymbol{\sigma} = [\sigma_1, \sigma_2, \ldots, \sigma_m] \in \mathbb{R}^m$ is the noise for each client, $\lambda > 0$ is a regularization parameter, and

$$L(\boldsymbol{B},\boldsymbol{\sigma}) = \sum_{i=1}^m l_i(\boldsymbol{\beta}_i, \sigma_i) \tag{4}$$

$$G(\boldsymbol{B},\boldsymbol{\sigma}) = \sum_{i\neq h}^m \Lambda_i(\boldsymbol{\beta}_i, \sigma_i) \tag{5}$$

where $l_i(\boldsymbol{\beta}_i, \sigma_i)$ is the loss function constructed using the data of client $i$, and $\Lambda_i(\boldsymbol{\beta}_i, \sigma_i)$ measures the similarity of client $i$ with other clients.

Usually, the loss of LLS regression is formulated using the negative log-likelihood function. According to Eq. (2), the loss function of client $i$ is described as $l_i(\boldsymbol{\beta}_i, \sigma_i) = -\sum_{j=1}^{n_i}\left(\frac{1}{\sigma_i} + f\left(\frac{y_{ij} - \boldsymbol{x}_{ij}^T \boldsymbol{\beta}_i}{\sigma_i}\right)\right)$. Assume $\epsilon$ follows SEV distribution, where $f(\epsilon) = \exp(\epsilon - \exp(\epsilon))$, thus, the local loss function of client $i$ can be written as: $l_i(\boldsymbol{\beta}_i, \sigma_i) = -\sum_{j=1}^{n_i}\left(\frac{1}{\sigma_i} + \exp\left(\left(\frac{y_{ij} - \boldsymbol{x}_{ij}^T \boldsymbol{\beta}_i}{\sigma_i}\right) - \exp\left(\left(\frac{y_{ij} - \boldsymbol{x}_{ij}^T \boldsymbol{\beta}_i}{\sigma_i}\right)\right)\right)\right)$. Implementing $l_i(\boldsymbol{\beta}_i, \sigma_i)$ into Eq. (4), the total loss function $L(B, \boldsymbol{\sigma})$ can be rewritten as:

$$L(\boldsymbol{B}, \boldsymbol{\sigma}) = -\sum_{i=1}^{m} \sum_{j=1}^{n_i} \left( \frac{1}{\sigma_i} + exp\left( \left(\frac{y_{ij} - x_{ij}^T \boldsymbol{\beta}_i}{\sigma_i}\right) \right) - exp\left( \left(\frac{y_{ij} - x_{ij}^T \boldsymbol{\beta}_i}{\sigma_i}\right) \right) \right) \qquad (6)$$

The regularization term, $G(\boldsymbol{B}, \boldsymbol{\sigma})$, in the objective function (i.e., Eq. (3)) denotes the discrepancy between model parameter sets across different clients. The difference is captured by incorporating a similarity function $A$, which quantifies the variation between any two clients. Define $A(\cdot)$ as a similarity function such that $A: [0, \infty) \to \mathbb{R}$ is a non-linear function that satisfies the following properties [31]:

(1) $A$ is increasing and concave on $[0, \infty)$ and $A(0) = 0$;
(2) $A$ is continuously differentiable on $(0, \infty)$; and
(3) The derivative of $A$, $A'(t)$ satisfies $\lim_{t \to 0^+} A'(t)$ is finite.

The similarity function $A(\|\cdot\|^2)$ measures the difference between two elements in a non-linear manner, which is often evaluated by Euclidean distance. Accordingly, $A(\|\boldsymbol{\beta}_i - \boldsymbol{\beta}_h\|^2)$ and $A(\|\sigma_i - \sigma_h\|^2)$ evaluate the similarity of $\boldsymbol{\beta}_i$ and $\boldsymbol{\beta}_h$, $\sigma_i$ and $\sigma_h$, respectively. Substituting the similarity function $A$ in Eq. (5), the regularization term $G(\boldsymbol{B}, \boldsymbol{\sigma})$ can be expressed as:

$$G(\boldsymbol{B}, \boldsymbol{\sigma}) = \sum_{i \neq h}^{m} A(\|\boldsymbol{\beta}_i - \boldsymbol{\beta}_h\|^2) + \sum_{i \neq h}^{m} A(\|\sigma_i - \sigma_h\|^2) \qquad (7)$$

Some kennel functions are commonly adopted as similarity functions, including negative exponential, smoothly clipped absolute deviation, and minimax concave penalty functions, which are summarized as follows [32]. We employ the widely used negative exponential function for our method.

(1) Negative exponential function: $A(\|\cdot\|^2) = 1 - \exp^{-\frac{\|\cdot\|^2}{\theta}}$

where $\theta > 0$.

(2) Smoothly clipped absolute deviation: $A(\|\cdot\|^2) = \begin{cases} \theta\|\cdot\|^2, & if \ \|\cdot\|^2 \leq \theta, \\ -\frac{\left((\|\cdot\|^2)^2 - 2\theta\lambda\|\cdot\|^2 + \tau^2\right)}{2(\alpha-1)\tau} & if \ \lambda < \|\cdot\|^2 \leq \theta\lambda, \\ C, & if \ \|\cdot\|^2 > \theta\lambda, \end{cases}$

where $\theta > 2$, $\lambda > 0$, and $C$ ensures continuity.

(3) Minimax concave penalty: $A(\|\cdot\|^2) = \begin{cases} \lambda\|\cdot\|^2 - \frac{(\|\cdot\|^2)^2}{2\theta}, & if \ \|\cdot\|^2 \leq \theta\lambda, \\ \frac{\theta\lambda^2}{2}, & if \|\cdot\|^2 > \theta\lambda, \end{cases}$

where $\lambda > 0$ and $\theta > 1$.

In summary, the first term $L(B, \boldsymbol{\sigma})$ represents the total loss function, which is the sum of the training loss of the local models of each client. This term allows each client to separately train its local model using its private training data. The second term $G(B, \boldsymbol{\sigma})$ is the regularization term, which consolidates the difference of model parameter sets by a similarity function. This component penalizes the model parameters with substantial deviations from those of other clients. The balance between $L(B, \boldsymbol{\sigma})$ and $G(B, \boldsymbol{\sigma})$ achieves an optimal trade-off between model personalization and generalization, ensuring reliability and adaptability to individual client-specific requirements.

By substituting Eqs. (6) and (7) into Eq. (3), the optimization formulation is complete. To make Eq. (3) convex, we apply a parameter transformation on $\boldsymbol{\beta}_i$ and $\sigma_i$, where $\widetilde{\boldsymbol{\beta}}_i = \boldsymbol{\beta}_i/\sigma_i$ and $\tilde{\sigma}_i = 1/\sigma_i$. Consequently, solving Eq. (4.1) is equivalent to solving the following:

$$\min_{\widetilde{\boldsymbol{B}},\widetilde{\boldsymbol{\sigma}}} \left( L(\widetilde{\boldsymbol{B}},\widetilde{\boldsymbol{\sigma}}) + \lambda G(\widetilde{\boldsymbol{B}},\widetilde{\boldsymbol{\sigma}}) \right) \tag{8}$$

where $L(\widetilde{\boldsymbol{B}},\widetilde{\boldsymbol{\sigma}}) = \sum_{i=1}^{m} l_i(\widetilde{\boldsymbol{\beta}}_i, \tilde{\sigma}_i) = -\sum_{i=1}^{m} \left( -n_i \log(\tilde{\sigma}_i) - \sum_{j=1}^{n_i}(y_{ij}\tilde{\sigma}_i - x_{ij}^T\widetilde{\boldsymbol{\beta}}_i) + \sum_{j=1}^{n_i} \exp\left((y_{ij}\tilde{\sigma}_i - x_{ij}^T\widetilde{\boldsymbol{\beta}}_i)\right) \right)$, and $G(\widetilde{\boldsymbol{B}},\widetilde{\boldsymbol{\sigma}}) = \sum_{i\neq h}^{m} A\left(\|\widetilde{\boldsymbol{\beta}}_i - \widetilde{\boldsymbol{\beta}}_h\|^2\right) + \sum_{i\neq h}^{m} A(\|\tilde{\sigma}_i - \tilde{\sigma}_h\|^2)$.

## 3. Federated Parameter Estimation Algorithm Based on PGD

Eq. (8) can be directly solved using convex optimization packages if the extracted features and TTFs of all clients (i.e., $x_{ij}$ and $y_{ij}$) could be shared or merged on the server, while such centralization is operationally impractical due to privacy concerns. Thus, we adopt an FL framework for model training to prevent data sharing with other clients or the central server. To handle the data heterogeneity issue, we further introduce a personalized FL approach, where each client maintains a personalized cloud model on the server instead of enforcing one single global model for all clients. This scheme results in a two-step optimization process for each client, which motivates us to propose a federated parameter estimation algorithm based on PGD to simultaneously realize solution personalization and privacy preservation. In the following paragraphs, we first outline the general steps of PGD, and then demonstrate the application of PGD in our algorithm, where the proposed algorithm is summarized.

PGD is an optimization method used to solve convex problems involving decomposable functions. It works by iteratively taking gradient steps to optimize one part of the objective, followed by a proximal operation on the other part. In our model, the objective function (i.e., Eq. (8)) can be decomposed into two parts, the loss function $L(\widetilde{\boldsymbol{B}},\widetilde{\boldsymbol{\sigma}})$ and the regularization term $G(\widetilde{\boldsymbol{B}},\widetilde{\boldsymbol{\sigma}})$. Based on incremental-type optimization [33], PGD solves this by alternatively optimizing $L(\widetilde{\boldsymbol{B}},\widetilde{\boldsymbol{\sigma}})$ and $G(\widetilde{\boldsymbol{B}},\widetilde{\boldsymbol{\sigma}})$ until convergence.

The first step is to optimize the regularization term $G(\widetilde{\boldsymbol{B}},\widetilde{\boldsymbol{\sigma}})$ using gradient descent. For simplicity, we define $\widetilde{\boldsymbol{W}} = \left[\widetilde{\boldsymbol{\beta}}^T, \widetilde{\boldsymbol{\sigma}}\right]^T \in \mathbb{R}^{(K+2)\times m}$, where $\widetilde{\boldsymbol{W}}$ denotes the parameter matrix of all clients. The updated $\tilde{S}$ in the $t^{th}$ iteration follows the update rule of the standard gradient descent, expressed as:

$$\widetilde{\boldsymbol{S}}^{(t)} = \widetilde{\boldsymbol{W}}^{(t-1)} - \alpha \nabla G\left(\widetilde{\boldsymbol{W}}^{(t-1)}\right) \tag{9}$$

where $\alpha$ is the learning rate that controls the step size, and $\nabla G(\widetilde{W}^{(t-1)})$ is the gradient of $G$ at $\widetilde{W}^{(t-1)}$ with the expression of:

$$\nabla G(\widetilde{W}^{(t-1)}) = \left(\frac{\partial G(\widetilde{W}^{(t-1)})}{\partial \widetilde{\sigma}}, \frac{\partial G(\widetilde{W}^{(t-1)})}{\partial \widetilde{\beta}_0}, \dots, \frac{\partial G(\widetilde{W}^{(t-1)})}{\partial \widetilde{\beta}_K}\right)^T \quad (10)$$

The second step is to optimize $L(\widetilde{B}, \widetilde{\sigma})$, equivalently, $L(\widetilde{W})$, where we employ the proximal mapping on $L(\widetilde{W})$ based on the result we obtain in the first step. This yields the updated $\widetilde{W}^{(t)}$ becoming:

$$\widetilde{W}^{(t)} = arg\min_{\widetilde{W}} L(\widetilde{W}) + \frac{\lambda}{2\alpha}\|\widetilde{W} - \widetilde{S}^{(t)}\|_2^2 \quad (11)$$

In summary, PGD consists of two key steps: gradient descent and proximal mapping. These steps can be applied in a federated manner, allowing for decentralized optimization across multiple clients. The detailed proof of its feasibility in a decentralized setting is shown in Appendix A. Thus, to solve Eq. (8) in a decentralized manner, equivalently, each client will optimize:

$$\min_{\widetilde{w}}(l_i(\widetilde{w}_i) + \lambda g_i(\widetilde{w}_i)) \quad (12)$$

where $\widetilde{w}_i = \left[\widetilde{\beta}_i^T, \widetilde{\sigma}_i\right]^T \in \mathbb{R}^{(K+2)}$, $l_i(\widetilde{w}_i) = n_i \log(\widetilde{\sigma}_i) + \sum_{j=1}^{n_i}(y_{ij}\widetilde{\sigma}_i - x_{ij}^T\widetilde{\beta}_i) - \sum_{j=1}^{n_i} exp\left((y_{ij}\widetilde{\sigma}_i - x_{ij}^T\widetilde{\beta}_i)\right)$, and $g_i(\widetilde{w}_i) = \sum_{h \neq i} A(\|\widetilde{w}_i - \widetilde{w}_h\|^2)$.

Specifically, in the first step, each client independently updates its parameters by optimizing the regularization term $g_i(\widetilde{w}_i)$ using gradient descent, described as:

$$\widetilde{s}_i^{(t)} = \widetilde{w}_i^{(t-1)} - \alpha \nabla g_i(\widetilde{w}_i^{(t-1)}) \quad (13)$$

In the second step, each client refines its model by optimizing its local loss function $l_i(\widetilde{w}_i)$ through proximal mapping, which is depicted as:

$$\widetilde{w}_i^{(t)} = arg\min_{\widetilde{w}_i} l_i(\widetilde{w}_i) + \frac{\lambda}{2\alpha}\|\widetilde{w}_i - \widetilde{s}_i^{(t)}\|_2^2 \quad (14)$$

The iterative process continues until convergence, ensuring that each client achieves an optimal balance between personalization and global regularization.

To perform personalized FL while preserving data privacy, we implement the optimization steps of PGD in a personalized client-server framework, where each client maintains a personalized cloud model on the server. Communication occurs through weighted model-aggregation messages between local models and the corresponding personalized cloud models. We first optimize $g_i(\widetilde{w}_i)$ on the server via gradient descent by Eq. (13). Here we treat $\widetilde{s}_i^{(t)}$ and $\widetilde{w}_i^{(t-1)}$ as the model parameter sets in the personalized cloud model and the local model of client $i$, respectively. Recall that $g_i(\widetilde{w}_i) = \sum_{h \neq i}^{m} A(\|\widetilde{w}_i - \widetilde{w}_h\|^2)$ and $A(\cdot)$ is a similarity function. Taking the first order derivative of $g_i(\widetilde{w}_i)$ with respect to $\widetilde{w}_i$, and substituting it into Eq. (13), the update $\widetilde{s}_i^{(t)}$ can be rewritten as Eq. (15). Detailed derivation can be found in Appendix B.

$$\tilde{s}_i^{(t)} = \left(1 - \gamma \sum_{h \neq i} A'\left(\left\|\widetilde{w}_i^{(k-1)} - \widetilde{w}_h^{(k-1)}\right\|^2\right)\right)\widetilde{w}_i^{(k-1)} + \gamma \sum_{h \neq i} A'\left(\left\|\widetilde{w}_i^{(k-1)} - \widetilde{w}_h^{(k-1)}\right\|^2\right)\widetilde{w}_h^{(k-1)}$$

$$= a_{i,1}\widetilde{w}_1^{(t-1)} + a_{i,2}\widetilde{w}_2^{(t-1)} + \cdots + a_{i,m}\widetilde{w}_m^{(t-1)} \qquad (15)$$

where $\gamma = 2\alpha$, $A'(\|\widetilde{w}_i - \widetilde{w}_h\|^2)$ is the derivative of $A(\|\widetilde{w}_i - \widetilde{w}_h\|^2)$, and $a_{i,1}, a_{i,2}, \ldots, a_{i,m}$ represent the linear combination weights corresponding with the model parameter sets $\widetilde{w}_1^{(t-1)}, \widetilde{w}_2^{(t-1)}, \ldots, \widetilde{w}_m^{(t-1)}$, respectively.

Since $a_{i,1} + a_{i,2} + \cdots + a_{i,m} = 1$, and $a_{i,1}, a_{i,2}, \ldots, a_{i,m}$ are non-negative, $\tilde{s}_i^{(t)}$ is a convex combination of the model parameter sets of local models, $\widetilde{w}_1^{(t-1)}, \widetilde{w}_2^{(t-1)}, \ldots, \widetilde{w}_m^{(t-1)}$. This can be interpreted as a weighted message aggregation mechanism depicted in Figure 1. For any client $i$, the parameter sets in the personalized cloud server, $\tilde{s}_i^{(t)}$, is a weighted aggregation of the messages received from the local models of all clients, $\widetilde{w}_1^{(t-1)}, \widetilde{w}_2^{(t-1)}, \ldots, \widetilde{w}_m^{(t-1)}$. The message from any client $h$, where $h \neq i$, is assigned a weight, $a_{i,h}$:

$$a_{i,h} = \gamma A'\left(\left\|\widetilde{w}_i^{(t-1)} - \widetilde{w}_h^{(t-1)}\right\|^2\right), (h \neq i) \qquad (16)$$

where $a_{i,h}$ is the contribution of $\widetilde{w}_h^{(t-1)}$ sent from client $h$ to the aggregated model parameter set $\tilde{s}_i^{(t)}$ of the personalized cloud model of client $i$, and $A'$ is the first derivative of similarity function $A$. The self-weight of client $i$, $a_{i,i}$, is the remaining portion excluding the weights from all other clients: $a_{i,i} = 1 - \gamma \sum_{h \neq i}^{m} A'\left(\left\|\widetilde{w}_i^{(t-1)} - \widetilde{w}_h^{(t-1)}\right\|^2\right)$.

The weight $a_{i,h}$ in Eq. (16) is determined by the function $A'$ and the Euclidean distance between clients $h$ and client $i$. As defined in Section 2, the similarity function $A$ is increasing and concave on $[0, \infty)$, implying its first derivative, $A'$, is a non-negative and non-increasing function on $(0, \infty)$. Consequently, a smaller Euclidean distance indicates higher similarity, which results in a larger weight $a_{i,h}$. When the model parameter sets of clients $i$ and $h$ (i.e., $\widetilde{w}_i^{(t-1)}$ and $\widetilde{w}_h^{(t-1)}$) are highly similar, their contributions to the updates in the personalized cloud server, $\tilde{s}_h^{(t)}$ and $\tilde{s}_i^{(t)}$, increase accordingly. This mechanism iteratively encourages collaboration among clients with similar model parameter sets, significantly enhancing the performance of personalized FL.

It is worth mentioning that the above communication process works without compromising data privacy. During the message-passing phase, each client transmits only its local model parameter sets, $\widetilde{w}_1^{(t-1)}, \widetilde{w}_2^{(t-1)}, \ldots, \widetilde{w}_m^{(t-1)}$, to the personalized cloud server. The cloud server then aggregates these parameters and returns a personalized model update, $\tilde{s}_i^{(t)}$, to each client. Such an exchange relies exclusively on model parameter sets, and each client does not even know the underlying message-passing graph. As a result, the process effectively optimizes $g_i(\widetilde{w}_i)$ without infringing the data privacy of all clients.

After optimizing $g_i(\widetilde{w}_i)$, each client receives an update $\tilde{s}_i$ from the personalized cloud server, which is subsequently used to optimize $l_i(\widetilde{w}_i)$. Each client will locally train its model by implementing the proximal operator using Eq. (14). Due to the parameter transformation, the optimization problem can be directly solved using convex optimization packages. The obtained model parameter set, $\widetilde{w}_i^{(t)}$, where $\widetilde{w}_i^{(t)} = \left[\widetilde{\beta}_i^{(t)}, \tilde{\sigma}_i^{(t)}\right]^T$, is the optimal solution at the $t^{th}$ iteration, which will be sent to the personalized cloud server for the next iteration. Since Eq. (14) depends solely on $l_i(\widetilde{w}_i)$ and $\tilde{s}_i^{(t)}$, the optimization is performed using the private training data of each client. Therefore, the whole process guarantees that the private training data is not exposed to other clients or the cloud server.

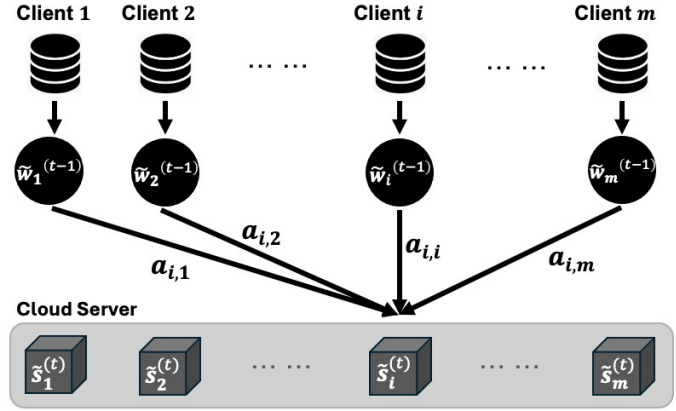

Figure 1. Weighted Message Aggregation Mechanism

The proposed personalized federated algorithm is summarized in Algorithm 1, which works as follows. In the first step, all the parameters on the clients, $\widetilde{w}_1^{(0)}, \dots, \widetilde{w}_m^{(0)}$, are initialized by sampling from a uniform distribution over the interval (0,10), i.e., $\widetilde{w}_i^{(0)} \sim U(0,10)^{K+2}$. In the second step, each client optimizes the objective function, Eq. (12), into two sub-steps: (i) The $g_i(\widetilde{w}_i)$ in Eq. (12) is first optimized by applying the gradient descent. Specifically, each personalized cloud server downloads the parameters $\widetilde{w}_1^{(0)}, \dots, \widetilde{w}_m^{(0)}$ from clients to compute $\tilde{s}_1^{(t)}, \dots, \tilde{s}_m^{(t)}$ by solving Eq. (3.13). The stored $\tilde{s}_1^{(t)}, \dots, \tilde{s}_m^{(t)}$ in each personalized cloud model is simply a weighted average of $\widetilde{w}_1^{(t-1)}, \dots, \widetilde{w}_m^{(t-1)}$. (ii) The $l_i(\widetilde{w}_i)$ in Eq. (12) is then optimized by applying a proximal operator on the update we obtain in (i). For client $i$, the server sends the model parameters in the personalized cloud model, $\tilde{s}_i^{(t)}$, to the client. Client $i$ is requested to perform local training with the knowledge of $\tilde{s}_i^{(t)}$. By solving Eq. (14), client $i$ computes an updated local model parameter $\widetilde{w}_i^{(t)}$.

Steps (i) and (ii) are repeated until the number of iterations reaches the maximum $M^*$. In the third step, each client downloads the optimal parameters $\widetilde{\boldsymbol{w}}_i^{(t)} = \left(\widetilde{\boldsymbol{\beta}}_i^{(t)^T}, \tilde{\sigma}_i^{(t)}\right)^T$, which is denoted as $\widetilde{\boldsymbol{w}}_i^* = \left(\widetilde{\boldsymbol{\beta}}_i^{*T}, \tilde{\sigma}_i^*\right)^T$.

---

**Algorithm 1**: Personalized Federated Parmeter Estimation

**Input**: Extracted Features $\left\{\{\boldsymbol{x}_{ij}\}_{j=1}^{n_i}\right\}_{i=1}^m \in \mathbb{R}^{K+1}$ and the corresponding TTFs $\left\{\{y_{ij}\}_{j=1}^{n_i}\right\}_{i=1}^m \in \mathbb{R}$, learning rate $\alpha$, and maximum iteration number $M^*$

**Output**: Optimal parameters $\{\widetilde{\boldsymbol{w}}_i^*\}_{i=1}^m \in \mathbb{R}^{(K+2) \times m}$, where $\widetilde{\boldsymbol{w}}_i^* = \left(\widetilde{\boldsymbol{\beta}}_i^{*T}, \tilde{\sigma}_i^*\right)^T$, and $\{\widetilde{\boldsymbol{s}}_i^*\}_{i=1}^m \in \mathbb{R}^{(K+2) \times m}$

(1) Randomly initialize parameters $\widetilde{\boldsymbol{w}}_1^{(0)}, \ldots, \widetilde{\boldsymbol{w}}_m^{(0)}$ on the clients, where $\widetilde{\boldsymbol{w}}_i^{(0)} = \left(\widetilde{\boldsymbol{\beta}}_i^{(0)^T}, \tilde{\sigma}_i^{(0)}\right)^T$, $\tilde{\sigma}_i^{(0)} \sim U(0,10)$, and $\widetilde{\boldsymbol{\beta}}_i^{(0)} \sim U(0,10)^{K+1}$

(2) For the number of iteration $t = 1, 2, \ldots, M^*$ do

    (i) Optimize $\tilde{g}_i(\widetilde{\boldsymbol{\beta}}_i, \tilde{\sigma}_i)$: Each personalized cloud server downloads $\widetilde{\boldsymbol{w}}_1^{(t-1)}, \ldots, \widetilde{\boldsymbol{w}}_m^{(t-1)}$ from the clients to compute $\widetilde{\boldsymbol{s}}_1^{(t)}, \ldots, \widetilde{\boldsymbol{s}}_m^{(t)}$ by solving Eq. (13).

    (ii) Optimize $\tilde{l}_i(\widetilde{\boldsymbol{\beta}}_i, \tilde{\sigma}_i)$: Client $i$, $i = 1, \ldots, m$, requests $\widetilde{\boldsymbol{s}}_1^{(t)}, \ldots, \widetilde{\boldsymbol{s}}_m^{(t)}$ from the personalized cloud server to compute $\widetilde{\boldsymbol{w}}_1^{(t)}, \ldots, \widetilde{\boldsymbol{w}}_m^{(t)}$ by solving Eq. (15).

    (iii) Repeat (i) and (ii) until the iteration number reaches $M^*$

(3) Client $i$, $i = 1, \ldots, m$, downloads $\widetilde{\boldsymbol{w}}_1^{(t)}, \ldots, \widetilde{\boldsymbol{w}}_m^{(t)}$, where $\widetilde{\boldsymbol{w}}_i^{(t)} = \left(\widetilde{\boldsymbol{\beta}}_i^{(t)^T}, \tilde{\sigma}_i^{(t)}\right)^T$

---

## 4. Simulation Study I

### *4.1 Data Generation and Benchmarks*

Assume that 10 clients jointly participate in establishing the proposed federated prognostic model, with each client possessing degradation data from 100 failed instances. For each client, 50 instances are randomly selected for model fitting, and the remaining 50 are used for testing. The input features and corresponding TTFs for each client are generated using the same sequential procedure. Specifically, we first generate the underlying degradation paths following the form $x_{ij}(t) = -\frac{c_{ij}}{\ln(t)}$, $0 \leq t < 1$, and $j = 1, \ldots, 50$. Here, the regression coefficients are generated from $c_{ij} \sim N(4, \sigma_{scenario}^2)$, where $\sigma_{scenario}$ is a parameter controlling the degree of heterogeneity across clients. A larger value of $\sigma_{scenario}$ indicates greater variation in the regression coefficients across clients, implying more diverse underlying relationships between the features and TTFs. In contrast, a smaller value of $\sigma_{scenario}$ implies that clients share more similar data distributions, indicating a lower degree of heterogeneity. Next, the TTF is determined by the time that the underlying degradation path $x_{ij}(t)$ reaches a threshold $D = 2$ plus some noise. In other words, the observed TTFs are computed as $y_{ij} = -\frac{c_{ij}}{D} + \varepsilon_{ij}$, where $\varepsilon_{ij}$ follows a standard Smallest Extreme Value (SEV) distribution, i.e., $\varepsilon_{ij} \sim SEV(0,1)$. Finally, we generate the observed degradation data as discrete-time observations of the underlying degradation paths. The observed degradation signal at time point $\tau_{ij}$ is generated from $x_{ij}(\tau_{ij}) = -\frac{c_{ij}}{\ln(\tau_{ij})} + \epsilon_{ij}(\tau_{ij})$, where $\tau_{ij} = 0.001, 0.002, \ldots, \left\lfloor \frac{y_{ij}}{0.001} \right\rfloor \times 0.001$ ($\lfloor \cdot \rfloor$ is the floor operator) and $\epsilon_{ij}(\tau_{ij}) \sim N(0, 0.05^2)$. The 50 testing samples for each client are generated following the same procedure described above. The model parameters, including the regularization parameter $\lambda$, the step size $\alpha$ in the gradient descent, and the parameter $\theta$ in the similarity function, are estimated through leave-one-out cross-validation.

To evaluate the effectiveness of our proposed approach, we compare our model, referred to as "Personalized Federated Learning Model (PFL)" against two benchmark models: (1) Local Model: In this method, each client trains its own model independently using its local data without sharing any information with other clients. Since each client possesses its own set of features and TTFs, they can perform the regression analysis of local TTFs against their respective local features individually. Here, we employ SEV regression to model the relationship between TTFs and input features to ensure consistency in the comparison across models. The regression coefficients for each client are estimated using maximum likelihood estimation. Since no information is shared among clients, this model serves as a baseline to measure the potential advantage offered by FL methods. However, the primary limitation is that it does not incorporate knowledge from other clients, which may not be ideal when some clients have a limited amount of training data. (2) Conventional Federated Learning Model (CFL): This model follows the standard FL paradigm, where clients collaboratively train a shared global model while preserving data privacy. In each iteration, clients first perform local training by optimizing a shared objective function, which is achieved by minimizing the loss function using gradient descent, before sending the updated model parameters to a central server [6]. The server then aggregates these updates to refine a global model, which is redistributed to all clients in the next iteration. Unlike the personalized approach we proposed, CFL assumes that a single global model is optimal for all clients, meaning all clients will ultimately share the same estimated regression coefficients, which may fail to capture client-specific variations.

### *4.2 Model Performance Comparison*

We analyze the impact of data heterogeneity at two different levels by varying the scale parameter $\sigma_{scenario}$, which controls the dispersion of regression coefficients across clients. We set $\sigma_{scenario} = 0.5$ to represent low heterogeneity, and $\sigma_{scenario} = 1$ as high heterogeneity. The generated training and testing samples as described in Section 4.1 remain the same for both heterogeneity levels. For each heterogeneity level, we conduct 20 random permutations of the data to evaluate the proposed method and the benchmark models. The model performance is assessed using the mean absolute percentage error (MAPE), defined as:

$$MAPE = \frac{|\text{Predcited TTF} - \text{True TTF}|}{\text{True TTF}} \times 100\% \quad (17)$$

When data heterogeneity is low ($\sigma_{scenario} = 0.5$), the performance of our proposed method and two benchmarks are shown in Figure 2. We can observe that PFL achieves the highest prediction accuracy with a median (and IQR) of 3.19% (1.98), followed by CFL with a median (and IQR) of 4.08% (2.34), and local training has the lowest accuracy with a median (and IQR) of 5.97% (5.62). This outcome is expected since PFL, a redesign of CFL, enhances collaboration among similar clients while still allowing for personalization. In a low-heterogeneity setting, client data distributions are fairly similar, meaning that most clients can effectively contribute to a shared learning process. FL settings, including CFL and PFL, will benefit greatly from this scenario due to the aggregated information from multiple clients. CFL enforces a single global model across all clients, which is well-suited for most clients by leveraging a larger pool of data. However, PFL further refines this process by allowing greater flexibility. Similar clients can share information more effectively, while minor differences among clients are still allowed through personalized updates. As a result, PFL achieves the highest prediction accuracy by balancing collaboration and personalization. On the other hand, local training performs worse than both CFL and PFL since it does not allow knowledge transfer across clients. Each client is restricted to its limited local dataset, thus, the model may be overfitting or underfitting, leading to less reliable predictions.

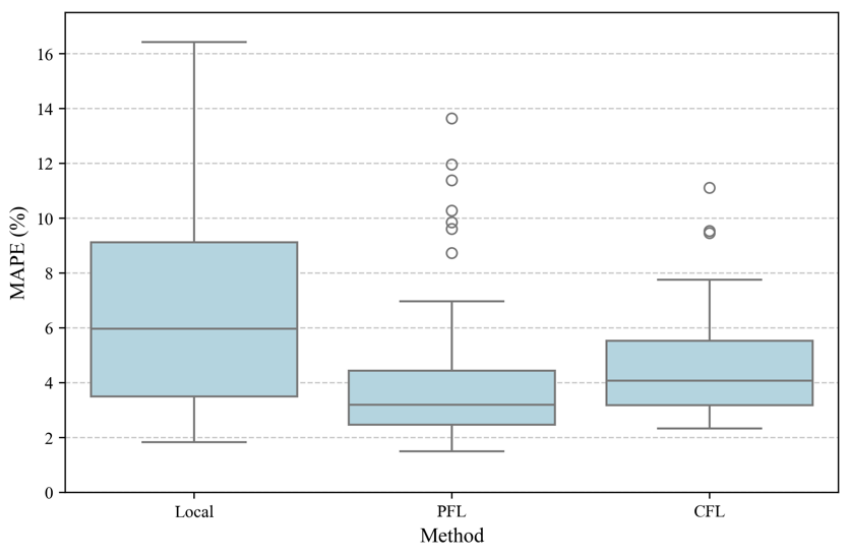

Figure 2. Prediction errors of the proposed method and two benchmarks under low data heterogeneity

As shown in Figure 3, when data heterogeneity is high ($\sigma_{scenario} = 1$), where there are significant variations among client data distributions, the ranking of prediction accuracy from the best to the worst is PFL with a median (and IQR) of 3.71% (1.40), local training with a median (and IQR) of 4.38% (2.97), and

CFL with a median (and IQR) of 5.48% (3.22). We can observe that PFL continues to perform the best. This is because PFL can identify similar clients, which enables them to benefit from collaborative learning while preventing interference from dissimilar clients. Thus, PFL effectively mitigates the impact of heterogeneity by adjusting model aggregation based on client similarity. In contrast, CFL struggles in high-heterogeneity settings since it forces all clients to share one single global model, which no longer generalizes well across diverse clients. The global model is negatively affected by the updates from diverse clients, leading to poor overall performance. For local training, it still suffers from the lack of inter-client collaboration, which prevents it from leveraging additional information to improve model accuracy. However, since each client trains independently, local training at least ensures that each model is tailored to its own dataset, avoiding the negative interference that occurs in CFL.

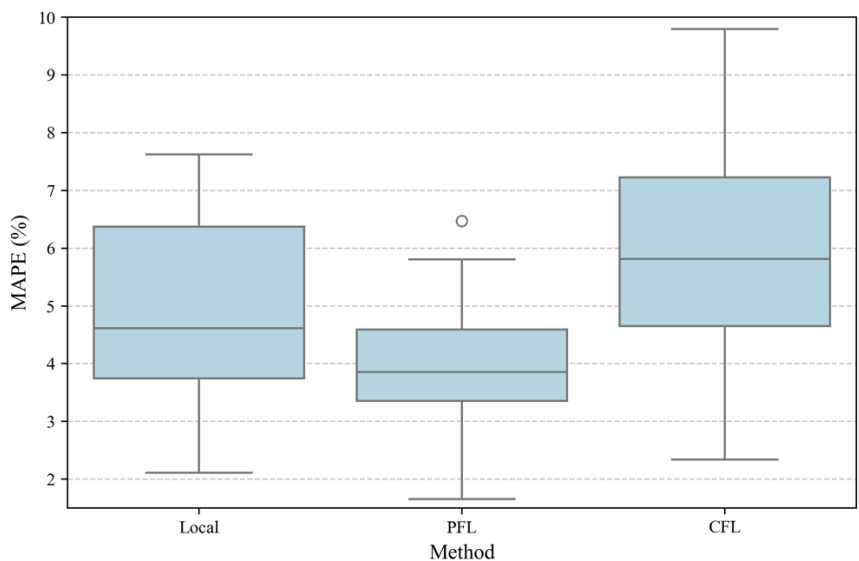

Figure 3. Prediction errors of the proposed method and two benchmarks under high data heterogeneity

Across different levels of heterogeneity, PFL consistently outperforms both CFL and local training by effectively balancing collaboration and personalization. When heterogeneity is low, federated learning methods (PFL and CFL) benefit from knowledge sharing, allowing clients to improve model accuracy through collaborative training. CFL performs well in this setting because the global model can generalize effectively across similar clients, but PFL further enhances performance by facilitating stronger collaboration among highly similar clients while still allowing for individualized model updates. In contrast, local training, which relies only on individual client data, performs the worst due to its inability to leverage information from other clients. As heterogeneity increases, the performance of CFL declines significantly because a single global model can no longer represent clients with diverse data distributions. The updates from various clients lead to poor generalization, making CFL the least effective approach in this scenario.

Local training appears to perform relatively better, while it is still constrained by data scarcity and the lack of collaborative learning. These results emphasize the importance of personalization in federated learning, making it possible to adapt to varying data distributions.

## 5. Simulation Study II

In this section, we investigate the impact of data quantity on model performance under both balanced and imbalanced data distributions. In the balanced scenario, each client owns the same number of samples. We will assess the influence of the variations in the total data volume on model performance. In the imbalanced case, where the number of samples differs among clients, we will examine how disparities in data quantity impact the performance of each client. In particular, we evaluate whether clients with fewer samples experience significant deviations in prediction accuracy compared to those with larger datasets. Similar as study I, the performance of our proposed model is compared with two benchmarks, CFL and local training models.

### *5.1 Model Performance under Balanced Data*

In the balanced data setting, we design multiple experimental scenarios, each characterized by a different total data volume. Assume there are 20 clients, and the data among clients are generated from the same underlying distribution, where $\sigma_{scenario} = 0.5$. The number of training samples per client increases from 5 to 15 across scenarios, with each client having the same number of samples. For example, each client has 5 samples in the first scenario, etc. The number of testing samples per client is fixed at 100 across all scenarios. The entire data generation and evaluation process is repeated 20 times. Figures 3-5 present the relation between the prediction error (i.e., MAPE) and the number of training samples per client of PFL, CFL, and local models, respectively. For the clarity of the comparison, we also report the median and IQR of MAPE for each model across different sample sizes, as shown in Table 1.

From Figures 4-6, we can observe that as the number of training samples per client increases from 5 to 15, the prediction error decreases consistently across all models. This trend is reasonable as larger sample sizes yield more stable and accurate estimates. A larger dataset can provide more information for parameter estimation, which leads to improved model stability and more reliable predictions. When clients have a limited number of training samples, the model is constrained by insufficient data, resulting in higher variability and greater uncertainty in prediction. This trend is particularly evident for local training, where the model solely relies on the data of an individual client. In contrast, FL methods, such as PFL and CFL, leverage shared information across clients, enabling better prediction even with limited data.

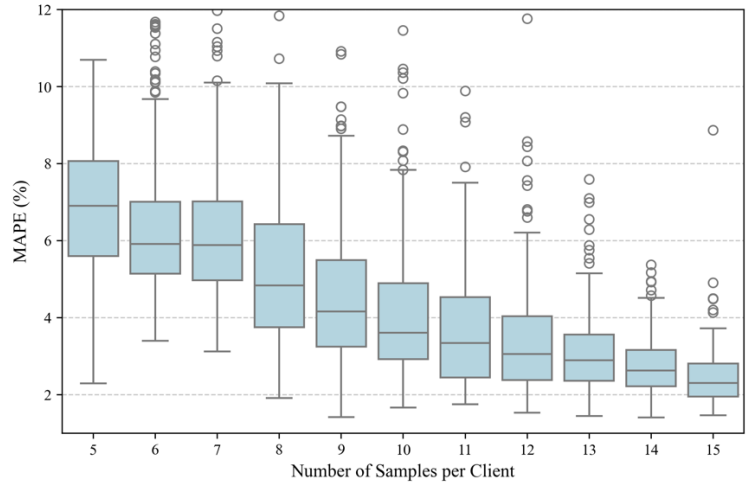

Figure 4. Relation between the prediction error of PFL and the number of training samples per client

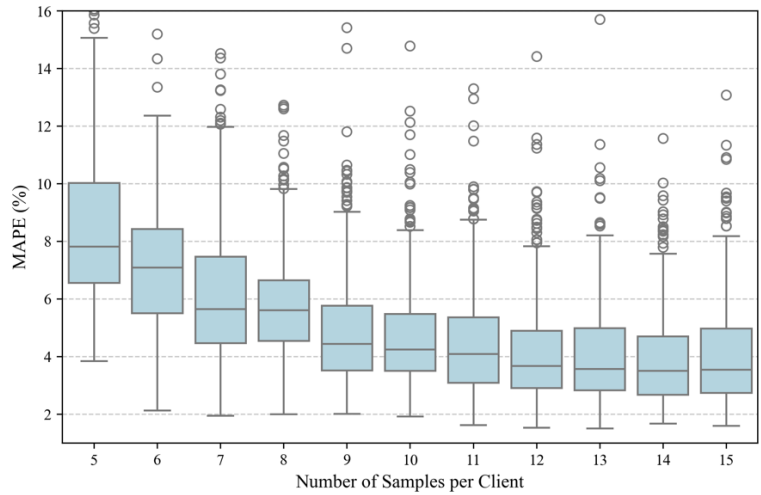

Figure 5. Relation between the prediction error of CFL and the number of training samples per client

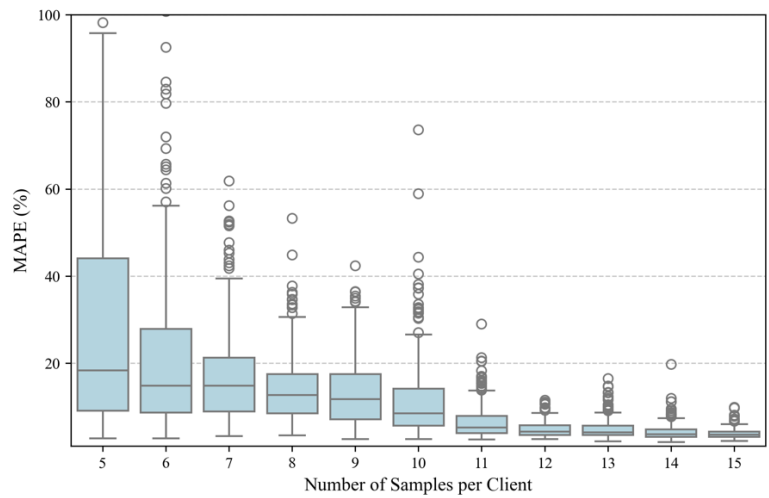

Figure 6. Relation between the prediction error of local training and the number of training samples per client

As shown in Table 1, when the number of training samples per client is relatively small (≤10), federated algorithms, including PFL and CFL, significantly outperform local training. For example, when each client has only five training samples, the median (and IQR) of MAPE for PFL and CFL are 6.91% (2.46) and 7.82% (3.48), respectively, whereas local training has a much higher median (and IQR) of 18.4% (34.9). This demonstrates that when the data is scarce, federated approaches enable clients to collaboratively train a model by leveraging information from other clients, reducing the negative impact of data scarcity. In contrast, local training relies exclusively on a client's own limited dataset, resulting in significantly higher prediction errors and greater variability. As the number of training samples per client exceeds 10, the performance of local training gradually improves, and becomes more comparable to FL methods, benefiting from the availability of more data per client. For instance, when each client has 15 training samples, local training achieves a median (and IQR) of 3.64% (1.21), approaching the performance of PFL and CFL, which are 2.31 (0.85) and 3.55 (2.23), respectively. However, FL still maintains an advantage in terms of prediction accuracy with the added benefit of data privacy. Comparing PFL and CFL, we observe that PFL consistently outperforms CFL. For example, when the number of training samples per client is 10, the median (and IQR) of PFL is 3.61 (1.97), which is lower than the median (and IQR) of CFL with 4.25 (1.97). This is because it customizes model updates based on the similarity of clients, while CFL assumes a single global model for all clients, which can be less effective in heterogeneous settings.

Table 1. Comparison of median (and IQR) between PFL, CFL, and local training across different total data amount

| Data amount per client | Median (and IQR) | | |
|---|---|---|---|
| | PFL | CFL | Local |
| 5 | 6.91 (2.46) | 7.82 (3.48) | 18.4 (34.90) |
| 6 | 5.92 (1.86) | 7.10 (2.93) | 14.9 (19.20) |
| 7 | 5.89 (2.06) | 5.65 (3.01) | 14.9 (12.30) |
| 8 | 4.84 (2.67) | 5.62 (2.11) | 12.8 (9.17) |
| 9 | 4.16 (2.25) | 4.44 (2.24) | 11.3 (10.10) |
| 10 | 3.61 (1.97) | 4.25 (1.97) | 8.50 (8.77) |
| 11 | 3.35 (2.08) | 4.10 (2.27) | 5.27 (4.03) |
| 12 | 3.05 (1.65) | 3.67 (1.98) | 4.36 (2.25) |
| 13 | 2.89 (1.20) | 3.58 (2.15) | 4.24 (2.29) |
| 14 | 2.63 (0.94) | 3.51 (2.02) | 3.74 (1.71) |
| 15 | 2.31 (0.85) | 3.55 (2.23) | 3.64 (1.21) |

## 5.2 Model Performance under Imbalanced Data

In real-world applications, data is often distributed unevenly across clients, leading to imbalanced learning scenarios. To evaluate the impact of such disparities on the prediction performance of PFL, we conduct experiments by considering 20 clients with varying amounts of training data. The data amount per client ranges from 50 to 240, increasing incrementally by 10 for each subsequent client. Compared with the scenario in Section 4.5.1, this setting introduces heterogeneity in data availability across clients. Each client is assigned 100 testing samples, and the entire training and evaluation process is replicated 30 times. Figure 4.6 presents the prediction error of PFL among participating clients with varying training sample sizes. We can interpret that the prediction error remains consistently around 4% for all clients despite the differences in data volume. This performance stability can be largely attributed to collaboratively learning among clients with similar characteristics. By leveraging shared knowledge, clients with smaller datasets benefit from insights through training with larger datasets, which effectively enhances predictive accuracy even with limited individual data. The adaptive learning mechanism ensures that predictive performance remains uniform across clients, mitigating the disparities that typically arise from differences in data availability.

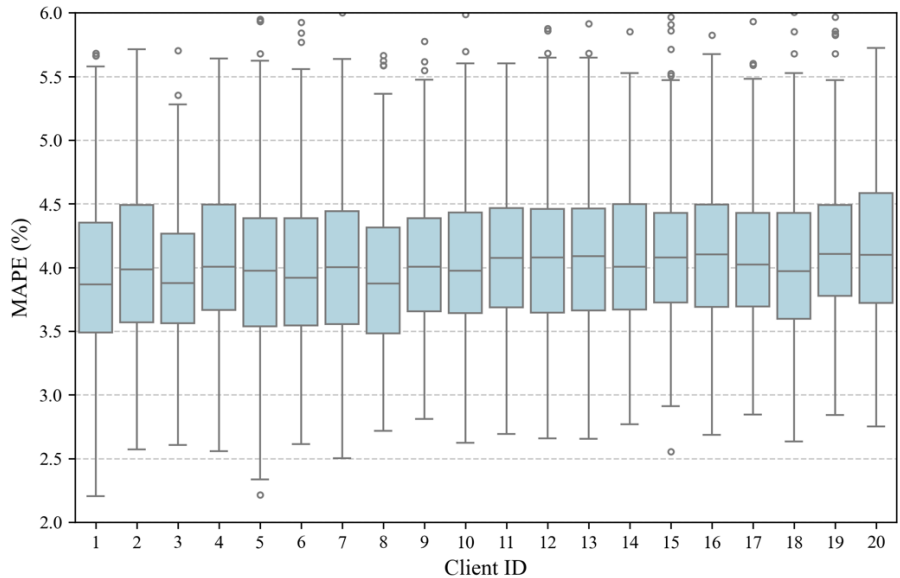

Figure 7. Prediction error of PFL among participating clients with varying training sample sizes

To further assess the effectiveness of our proposed model in scenarios with imbalanced data, we compare the performance of PFL against the local model trained individually for each client. Assume three clients with different training sample sizes. A total of 40 training samples are randomly distributed among the three clients, where Client 1, Client 2, and Client 3 possess 25, 10, and 5 samples, respectively. Each client is assigned 50 testing samples, and the training and evaluation process is repeated 50 times. Figures 7-9 present the comparison of prediction error between PFL and local training across Client 1, Client 2, and Client 3, respectively. The key observation is that PFL consistently provides more stable and accurate predictions than local training, regardless of the number of training samples available.

From Figure 8, the predictive accuracy of PFL and local training for Client 1 is comparable, with median (and IQR) of 0.115 (0.0253) and 0.117 (0.0296), respectively. This indicates that when a client has sufficient data, participating in PFL does not degrade its predictive performance, but it also provides only limited additional benefit. For Client 2 and Client 3, PFL provides substantial improvements in performance over local training, as shown in Figures 9 and 10. The median (and IQR) of MAPE for local training is 0.411 (0.548) for Client 2, and 27.8 (30.6) for Client 3, whereas PFL significantly reduces to 0.288 (0.192) and 0.521 (0.456), respectively. The considerable gap in performance highlights two key limitations of local training: (1) Inadequate generalization due to small sample sizes, and (2) High sensitivity to data availability. PFL mitigates these issues by enabling clients with insufficient training data to leverage information from other participants, allowing them to achieve more reliable and accurate predictions. This collaborative learning process guarantees that even clients with minimal data availability can benefit from other clients with adequate data without compromising privacy.

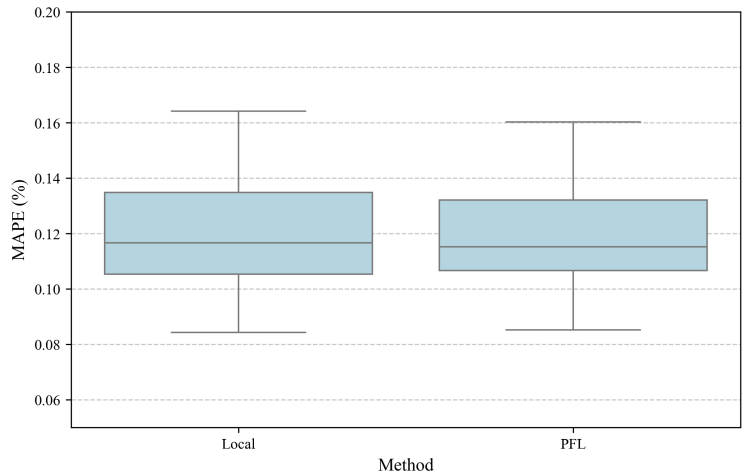

Figure 8. Prediction error of PFL and local training for Client 1

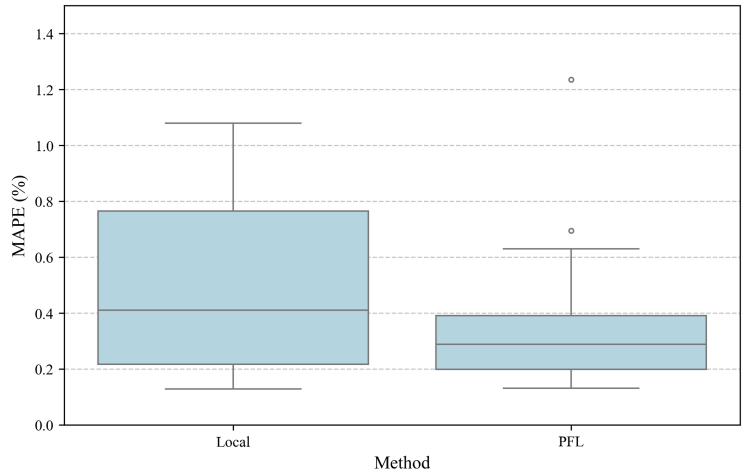

Figure 9. Prediction error of PFL and local training for Client 2

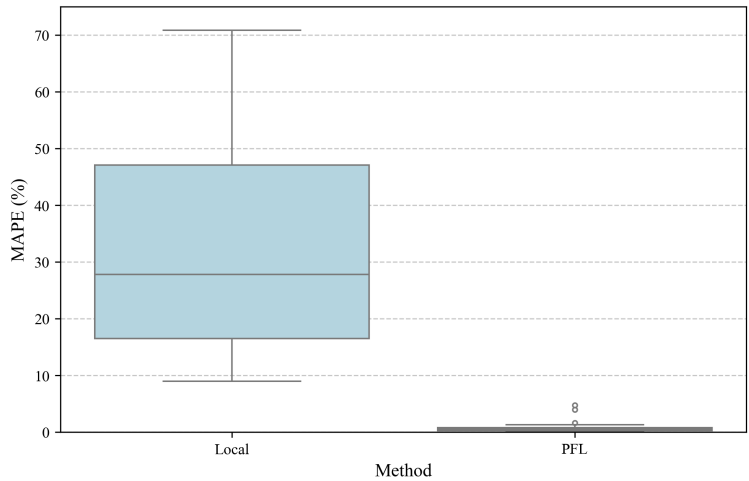

Figure 10. Prediction error of PFL and local training for Client 3

## 6. Case Study

In the aerospace industry, accurately predicting the RUL of aircraft engines is a critical task since it can help reduce operational disruptions and support cost-effective maintenance planning. One of the major challenges of prognostic model development lies in the limited availability of failure data in a single company. This is because engine failures are rare events due to high engineering standards and maintenance protocols. Although failure data can also be generated through accelerated degradation testing, the process is highly costly and time-consuming. Besides the limited data availability, many aerospace companies also face data privacy concerns. They are reluctant to share degradation data with external parties as the information is commercially sensitive. Therefore, it is beneficial for multiple companies to collaboratively develop prognostics models in a privacy-preserving manner such as FL approaches. However, conventional FL approaches face another key challenge, which is the data heterogeneity issue across companies. While engines may belong to the same manufacturer or model type, they often operate under various real-world conditions, including different environmental factors, maintenance practices, and operating cycles. In addition, engines may be monitored using different sensor types, configurations, or data acquisition protocols, resulting in inconsistent degradation patterns. These differences lead to data heterogeneity across companies, which can significantly degrade the performance of the global FL model. In this study, we apply our proposed PFL algorithm to a turban engine degradation dataset sourced from the NASA data repository to validate its effectiveness in handling such distributed and heterogeneous data.

## 6.1 Dataset Description and Preprocessing

The dataset used in this study is the multi-sensor degradation data from a fleet of aircraft turbofan engines with two failure modes [34]. It includes the degradation signals from 100 engines along with their failure times. Each engine is monitored by 21 sensors, and every sensor records time-series signals that capture the degradation process over operational time. Based on the study [1], not all sensors are informative for prognostics. Sensors 4, 15, 17, and 20 have been identified as the most informative, corresponding to critical physical measurements: total temperature at the low-pressure turbine outlet (Sensor 4), bypass ratio (Sensor 15), bleed enthalpy (Sensor 17), and high-pressure turbine coolant bleed (Sensor 20). Thus, in this study, we only use the degradation signals from these 4 sensors for prognostics.

According to sensor signal patterns, engines are classified into two failure modes, High-Pressure Compressor (HPC) degradation, referred to as Failure Mode 1 (FM1), and Fan degradation, represented as Failure Mode 2 (FM2). Among 100 engines, 56 engines belong to FM1, and 44 engines are labeled as FM2. Although the degradation processes of FM1 and FM2 are not identical, their degradation signals exhibit a high degree of correlation. This indicates that the two failure modes share common underlying degradation patterns, which motivates us to use our proposed prognostic model. To reflect inter-client heterogeneity, we assume engines from different failure modes are allocated to different subsets of clients. Assume 4 clients jointly participating in the development of prognostic models. Engines from FM1 are distributed evenly across Client 1 and Client 2, while engines from FM2 are evenly assigned to Client 3 and Client 4. For each client, 40% of the assigned engines are randomly selected for training, and the remaining 60% are used for testing. For a reliable performance assessment under each failure mode, the testing engines from FM1 are aggregated as the testing set for Client 1 and Client 2. Similarly, the testing engines from FM2 are combined as the testing set for Client 3 and Client 4. Noting that we truncate the testing signals to 70% of their whole length to simulate the case when the engine is prematurely terminated at random time points prior to their failure time, so that we can predict the number of remaining operational cycles before failure in the test set.

The multi-stream degradation signals are high-dimensional, which implies that the signals usually contain redundant information that could be fused to reduce the number of features used for constructing prognostic models. Thus, following the above partition, we further apply data fusion to the dataset owned by each client to fuse high-dimensional data and provide low-dimensional features. In this paper, we use smoothing spline for data fusion. Smoothing spline is a non-parametric regression that fits a smooth curve through time-series data by balancing goodness-of-fit and smoothness [35]. Cubic smoothing splines are adopted in this study. The penalty parameter controlling the smoothness is selected via cross-validation. For each client, we apply the smoothing spline on each sensor signal, where the basis of the same sensor remains the same. The TTFs are then mapped to the features extracted by splines using the statistical learning model. We compare the proposed model, PFL, with two benchmarks, CFL and local training model, which are described in the simulation study in Section 4. The proposed model and benchmarks use the same test dataset. The whole procedure including data partition, data fusion, and prognostic model construction is replicated 30 times.

## 6.2 Results and Analysis

The prediction performances of the proposed model and two benchmarks for Client 1, Client 2, Client 3, and Client 4 are reported in Figures 11-14, respectively. The figures show that the proposed method performs better than the two benchmarks for all clients. For example, the medians (and IQRs) of the proposed method, CFL, and local training model constructed by Client 1 are 0.197 (0.138), 0.485 (0.194), and 0.445 (0.502), respectively; for client 2, the medians (and IQRs) are 0.159 (0.152), 0.294 (0.115), and 0.291 (0.483), respectively; for client 3, the medians (and IQRs) are 0.157 (0.0688), 0.282 (0.0999), and 0.582 (0.157), respectively; and for client 4, the medians (and IQRs) are 0.155 (0.0694), 0.411 (0.0770), and 0.394 (0.303), respectively. The comparison results align with the simulation outcomes. This is expected since federated models (i.e., PFL and CFL) use training data from all 4 clients, while local training only uses its individual data for model training. Thus, the sample size of federated learning models is larger than that of each client. Since the model constructed using PFL can encourage similar clients to contribute more with each other, the predictive performance of PFL is better than that of CFL, which utilizes one global model for all clients. As a result, the model using PFL is more accurate than the one using CFL and the individual models trained by each client.

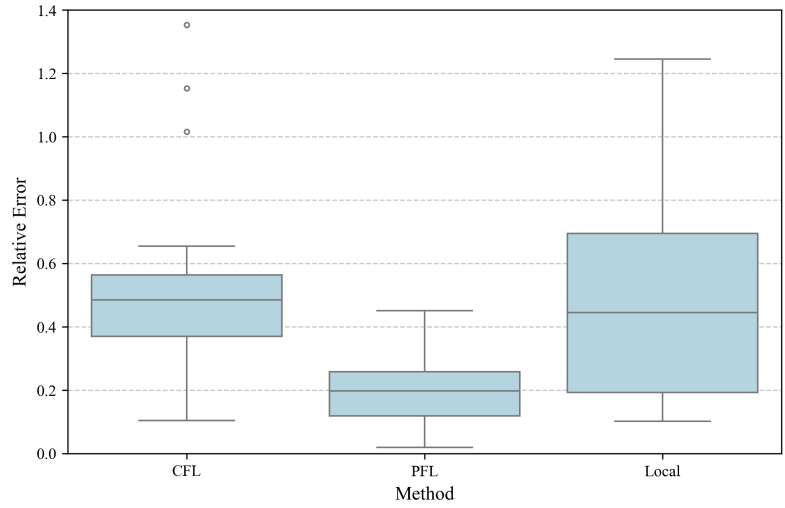

Figure 11. Prediction error of the proposed method and two benchmarks for Client 1

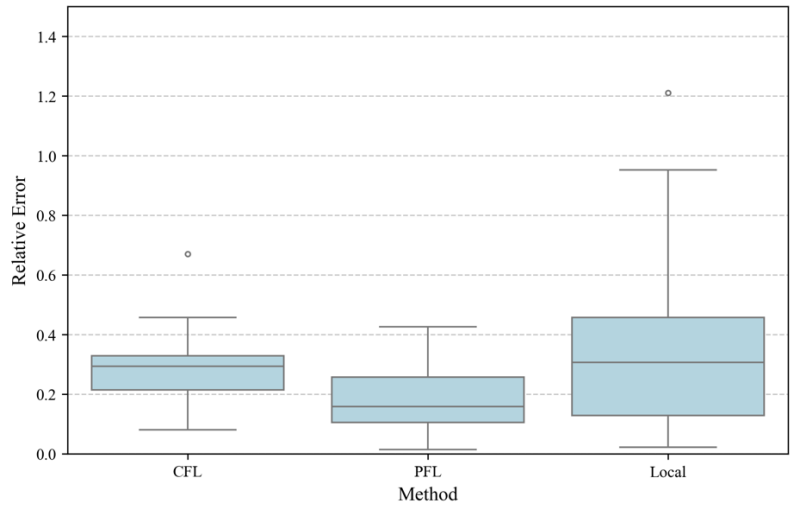

Figure 12. Prediction error of the proposed method and two benchmarks for Client 2

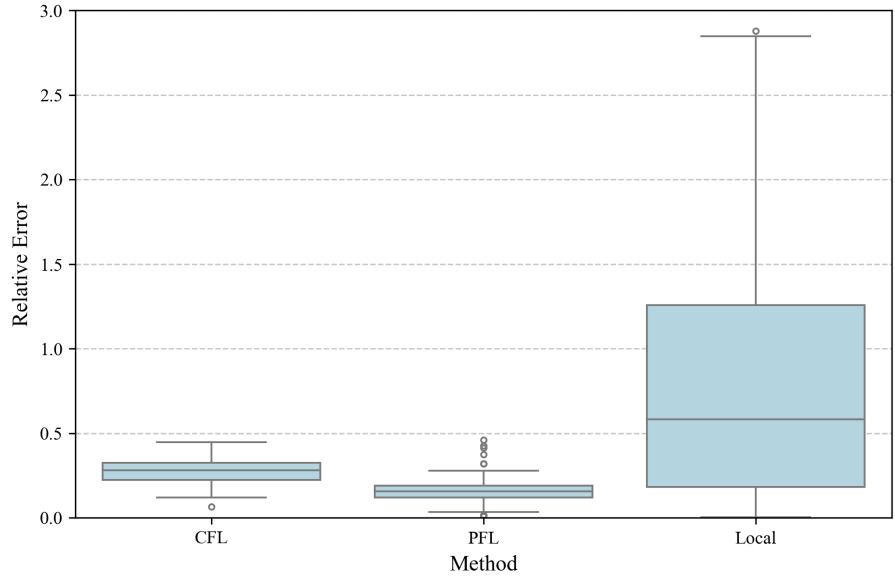

Figure 13. Prediction error of the proposed method and two benchmarks for Client 3

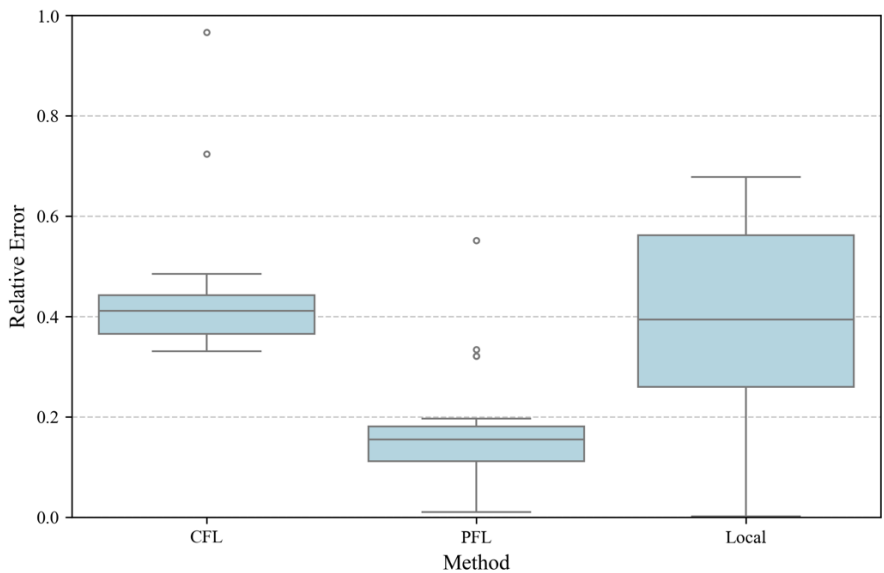

Figure 14. Prediction error of the proposed method and two benchmarks for Client 4

## 7. Conclusions

This paper introduces a personalized federated prognostic framework tailored for settings where clients exhibit heterogeneous degradation behaviors. Unlike traditional FL models that assume uniformity across clients, our approach allows each client to build a customized prognostic model while still benefiting from collaborative learning. By promoting selective information sharing among clients with similar degradation patterns, the model facilitates pairwise collaborations that enhance the model performance. To estimate parameters while keeping the data confidential, we further propose a federated parameter estimation algorithm based on proximal gradient descent. The algorithm alternates between two coordinated steps: a collaborative gradient update that encourages parameter alignment among similar clients, and a local refinement step using proximal mapping based on private client data. We have proved that these steps can be applied in a federated manner, allowing for decentralized optimization across multiple clients.

Our approach offers several key advantages over other alternatives by simultaneously realizing model personalization, preserving data privacy, and providing comprehensive failure time distributions, especially in scenarios with limited data and constrained resources. Comprehensive validation on both simulated datasets and the NASA turbofan engine degradation dataset demonstrates the effectiveness of the proposed approach. The results have shown an improved performance over local training models and traditional FL methods, under both low and high heterogeneity. Notably, clients with fewer training samples benefit significantly from the personalized structure of our model. By enabling knowledge transfer from clients with richer data, the approach helps mitigate performance disparities that commonly arise due to data imbalance, ultimately leading to more equitable and reliable prognostic performance across the client network.

# 8. Appendix

## 8.1 Appendix A

The $G(\widetilde{B}, \widetilde{\sigma})$ in Eq. (8), equivalently, $G(\widetilde{W})$, can be rewritten as $G(\widetilde{W}) = \sum_{i \neq h}^{m} A(\|\widetilde{w}_i - \widetilde{w}_h\|^2)$, where $\widetilde{w}_i$ and $\widetilde{w}_h$ are the parameter vectors for clients $i$ and $h$, respectively. Extracting the $i^{th}$ column of $G(\widetilde{W})$, the regularization term for client $i$, $g_i(\widetilde{w}_i)$, is described as: $g_i(\widetilde{w}_i) = \sum_{h \neq i} A(\|\widetilde{w}_i - \widetilde{w}_h\|^2)$. Thus, $G(\widetilde{W})$ is the sum of all client-specific terms $g_i(\widetilde{w}_i)$, represented as $G(\widetilde{W}) = \sum_{i=1}^{m} g_i(\widetilde{w}_i)$.

Similarly, the gradient of $g_i(\widetilde{w}_i)$ with respect to $\widetilde{w}_i$ in the $(t-1)^{th}$ iteration is computed as:

$$\nabla g_i\left(\widetilde{w}_i^{(t-1)}\right) = \left(\frac{\partial G_i\left(\widetilde{w}_i^{(t-1)}\right)}{\partial \widetilde{\sigma}}, \frac{\partial G_i\left(\widetilde{w}_i^{(t-1)}\right)}{\partial \widetilde{\beta}_0}, \ldots, \frac{\partial G_i\left(\widetilde{w}_i^{(t-1)}\right)}{\partial \widetilde{\beta}_K}\right)^T \quad (A.1)$$

Based on Eqs. (10) and (A.1), the gradient of $G(\widetilde{W})$ is the sum of the gradients of $g_i(\widetilde{w}_i)$, where $\nabla G(\widetilde{W}^{(t-1)}) = \sum_{i=1}^{m} \nabla g_i(\widetilde{w}_i^{(t-1)})$. Thus, Eq. (10) can be rewritten as:

$$\tilde{S}^{(t)} = \widetilde{W}^{(t-1)} - \alpha \sum_{i=1}^{m} \nabla g_i\left(\widetilde{w}_i^{(t-1)}\right) \quad (A.2)$$

This implies that in the first step of PGD, the total gradient is the sum of the gradients computed individually by each client. In other words, this process can be performed in a decentralized manner, where each client performs gradient descent individually. Thus, any client $i$ will update its parameters based on Eq. (13), which will be transmitted back to each client for the second step.

As illustrated in Eq. (4), the total loss $L(\widetilde{W})$ is the sum of loss functions of the individual client, i.e., $L(\widetilde{W}) = \sum_{i=1}^{m} l_i(\widetilde{w}_i)$. Since $\widetilde{W} = [\widetilde{w}_1, \widetilde{w}_2, \ldots, \widetilde{w}_m]$ and $\tilde{S}^{(t)} = [\tilde{s}_1^{(t)}, \tilde{s}_2^{(t)}, \ldots, \tilde{s}_m^{(t)}]$, the squared Euclidean norm $\|\widetilde{W} - \tilde{S}^{(t)}\|_2^2$ in Eq. (11) can be decomposed as: $\|\widetilde{W} - \tilde{S}^{(t)}\|_2^2 = \sum_{i=1}^{m} \|\widetilde{w}_i - \tilde{s}_i^{(t)}\|_2^2$. Thus, Eq. (11) becomes: $\widetilde{W}^{(t)} = \arg\min_{W} \sum_{i=1}^{m} \left(l_i(\widetilde{w}_i) + \frac{\lambda}{2\alpha} \|\widetilde{w}_i - \tilde{s}_i^{(t)}\|_2^2\right)$. This demonstrates that the proximal operation can be reformulated as client-specific subproblems, allowing each client to compute its own proximal update locally, where each client $i$ solves based on Eq. (14).

## 8.2 Appendix B

$$\tilde{s}_i^{(k)} = \widetilde{w}_i^{(k-1)} - \alpha \nabla \tilde{G}\left(\widetilde{w}_i^{(k-1)}\right)$$

$$= \widetilde{w}_i^{(k-1)} - \gamma \sum_{h \neq i} A'\left(\left\|\widetilde{w}_i^{(k-1)} - \widetilde{w}_h^{(k-1)}\right\|^2\right) \left(\left\|\widetilde{w}_i^{(k-1)} - \widetilde{w}_h^{(k-1)}\right\|\right)$$

$$= \widetilde{w}_i^{(k-1)} - \gamma \sum_{h \neq i} A'\left(\left\|\widetilde{w}_i^{(k-1)} - \widetilde{w}_h^{(k-1)}\right\|^2\right) \widetilde{w}_i^{(k-1)} + \gamma \sum_{h \neq i} A'\left(\left\|\widetilde{w}_i^{(k-1)} - \widetilde{w}_h^{(k-1)}\right\|^2\right) \widetilde{w}_h^{(k-1)}$$

$$= \left(1 - \gamma \sum_{h \neq i} A'\left(\left\|\widetilde{\boldsymbol{w}}_i^{(k-1)} - \widetilde{\boldsymbol{w}}_h^{(k-1)}\right\|^2\right)\right) \widetilde{\boldsymbol{w}}_i^{(k-1)} + \gamma \sum_{h \neq i} A'\left(\left\|\widetilde{\boldsymbol{w}}_i^{(k-1)} - \widetilde{\boldsymbol{w}}_h^{(k-1)}\right\|^2\right) \widetilde{\boldsymbol{w}}_h^{(k-1)}$$

$$= a_{i,1} \widetilde{\boldsymbol{w}}_1^{(k-1)} + a_{i,2} \widetilde{\boldsymbol{w}}_2^{(k-1)} + \cdots + a_{i,m} \widetilde{\boldsymbol{w}}_m^{(k-1)}$$

where $\gamma = 2\alpha$.

## 9. Data Availability Statement

The data that support the findings of this study are openly available in NASA Prognostics Center of Excellence Data Set Repository at https://www.nasa.gov/content/prognosticscenter-of-excellence-data-set-repository.